%% file: 0-main.tex
\documentclass{ieeeaccess}
\usepackage{cite}
\usepackage{amsmath,amssymb,amsfonts}
\usepackage{algorithmic}
\usepackage{graphicx}
\usepackage{textcomp}
\usepackage{booktabs}
\usepackage{tabularx}

\def\BibTeX{{\rm B\kern-.05em{\sc i\kern-.025em b}\kern-.08em
    T\kern-.1667em\lower.7ex\hbox{E}\kern-.125emX}}

\usepackage{xspace}

\makeatletter
\DeclareRobustCommand\onedot{\futurelet\@let@token\@onedot}
\def\@onedot{\ifx\@let@token.\else.\null\fi\xspace}

\def\eg{\emph{e.g}\onedot} 
\def\ie{\emph{i.e}\onedot} 
 
\def\etc{\emph{etc}\onedot} 
\def\wrt{w.r.t\onedot} 

\makeatother

\def\tinyml{TinyML\xspace}

\begin{document}


\history{Date of current version September 26, 2023.}
\doi{XXXXXXX.XXXXXXX}

\title{
A Machine Learning-oriented Survey on Tiny Machine Learning
}

\author{
\uppercase{Luigi Capogrosso}, \IEEEmembership{Student, IEEE},
\uppercase{Federico Cunico}, 
\uppercase{Dong Seon Cheng},\\
\uppercase{Franco Fummi}, \IEEEmembership{Member, IEEE},
\uppercase{Marco Cristani}, \IEEEmembership{Member, IEEE},
}
\address[]{Department of Engineering for Innovation Medicine, University of Verona, Verona, Italy, (e-mail: name.surname@univr.it)}

\tfootnote{
This study was carried out within the PNRR research activities of the consortium iNEST (Interconnected North-Est Innovation Ecosystem) funded by the European Union Next-GenerationEU (Piano Nazionale di Ripresa e Resilienza (PNRR) – Missione 4 Componente 2, Investimento 1.5 – D.D. 1058  23/06/2022, ECS\_00000043). This manuscript reflects only the Authors’ views and opinions, neither the European Union nor the European Commission can be considered responsible for them.
}

\markboth
{Capogrosso \headeretal: A Machine Learning-oriented Survey on Tiny Machine Learning}
{Capogrosso \headeretal: A Machine Learning-oriented Survey on Tiny Machine Learning}

\corresp{Corresponding author: Luigi Capogrosso (e-mail: luigi.capogrosso@univr.it).}

\begin{abstract}
The emergence of Tiny Machine Learning (\tinyml{}) has positively revolutionized the field of Artificial Intelligence by promoting the joint design of resource-constrained IoT hardware devices and their learning-based software architectures. 
\tinyml{} carries an essential role within the fourth and fifth industrial revolutions in helping societies, economies, and individuals employ effective AI-infused computing technologies (\eg{}, smart cities, automotive, and medical robotics). 
Given its multidisciplinary nature, the field of \tinyml{} has been approached from many different angles: this comprehensive survey wishes to provide an up-to-date overview focused on all the learning algorithms within \tinyml{}-based solutions. 
The survey is based on the Preferred Reporting Items for Systematic Reviews and Meta-Analyses (PRISMA) methodological flow, allowing for a systematic and complete literature survey. 
In particular, firstly we will examine the three different workflows for implementing a \tinyml{}-based system, \ie{}, ML-oriented, HW-oriented, and co-design. 
Secondly, we propose a taxonomy that covers the learning panorama under the \tinyml{} lens, examining in detail the different families of model optimization and design, as well as the state-of-the-art learning techniques. 
Thirdly, this survey will present the distinct features of hardware devices and software tools that represent the current state-of-the-art for \tinyml{} intelligent edge applications. 
Finally, we discuss the challenges and future directions. 
\end{abstract}


\titlepgskip=-21pt

\maketitle

\input{srcs/1_intro}
\input{srcs/2_existing_surveys}
\input{srcs/3_selection_criteria}
\input{srcs/4_tinyML_pipelines}
\input{srcs/5_efficient_dl}
\input{srcs/6_devices_and_tools}
\input{srcs/7_discussion}
\input{srcs/8_conclusion.tex}

\bibliographystyle{IEEEtran}
\bibliography{0.1-bibliography}


\begin{IEEEbiography}[{\includegraphics[width=1in,height=1.25in,clip,keepaspectratio]{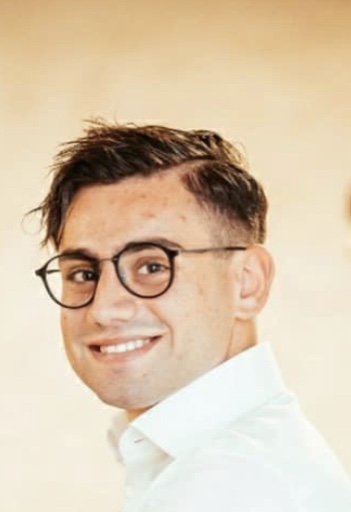}}]{Luigi Capogrosso} (Student, IEEE)
is a Ph.D. Student of the National Program in Artificial Intelligence at the University of Verona in collaboration with the Polytechnic of Turin in the IntelliGO labs under the supervision of Prof. Marco Cristani and Prof. Franco Fummi. 
He received his B.Sc. (2019) and M.Sc. (2021) at the University of Verona in Computer Science and Computer Engineering for Robotics and Smart Industry, respectively. 
His research interest can be grouped into the two areas of efficient deep learning and representation learning. 
\end{IEEEbiography}

\begin{IEEEbiography}[{\includegraphics[width=1in,height=1.25in,clip,keepaspectratio]{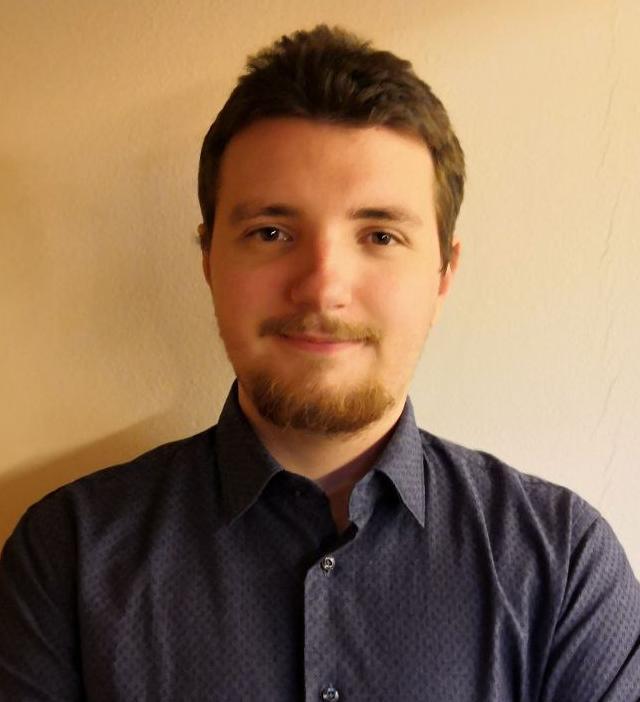}}]{Federico Cunico}
is a Ph.D. Candidate in Computer Science at the University of Verona under the supervision of Prof. Marco Cristani. 
He received his Master’s degree in Computer Science and Engineering at the University of Verona in 2019. 
His main research interests include deep learning and computer vision for industrial scene analysis, with emphasis on human-centered tasks such as attention estimation, human pose and motion forecasting, and human pose estimation. 
\end{IEEEbiography}

\begin{IEEEbiography}[{\includegraphics[width=1in,height=1.25in,clip,keepaspectratio]{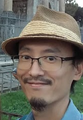}}]{Dong Seon Cheng}
is a Research Associate at the University of Verona from 2023.  
He received his Laurea degree (2005) and Ph.D. in Computer Science (2008) at the University of Verona with work in the field of computer vision and pattern recognition. 
Machine learning is his main interest, and he has published several conferences and journal papers in the field. 
From 2012 to 2017, he was an Assistant Professor at the Department of Computer Science and Engineering of the Hankuk University of Foreign Studies (South Korea) teaching undergraduate and graduate courses. 
From 2019 to 2022 he worked at SETECNA EPC S.rl., an electronics company in the HVAC industry. 
\end{IEEEbiography}

\begin{IEEEbiography}[{\includegraphics[width=1in,height=1.25in,clip,keepaspectratio]{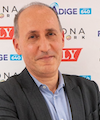}}]{Franco Fummi} (Member, IEEE)
received the Laurea degree in Electronic Engineering at the Polytechnic of Milan in 1990 and the Ph.D.in Electronic and Communication Engineering in 1994 at the Polytechnic of Milan. 
In 1993 he was a Research Assistant at the Department of Computer Science of the University of Victoria. 
In 1996 he obtained the position of Assistant Professor in Computer Science at the ``Dipartimento di Elettronica e Informazione'' of Polytechnic of Milan where he remained until October 1998. 
In July 1998 he obtained the position of Associate Professor in Computer Architecture at the Computer Science Department of Università di Verona. 
Since March 2001 he is a Full Professor in Computer Architecture at Università di Verona, before at the Computer Science Department and then at the Department of Engineering for Innovation Medicine. 
He is leading the Cyber-physical and IoT Systems Design (CISD) group of the Università di Verona, currently composed of more than 20 people, and working on hardware description languages and electronic design automation methodologies for modeling, verification, testing, and optimization of cyber-physical systems.  
Since 2018 he is the project manager of the Industrial Computer Engineering (ICE) Laboratory at the University of Verona: a facility serving as a technological demonstrator and research laboratory, functional to rethink industrial processes such as additive and subtractive manufacturing, quality control, assembly, and parts storage. 
He is also a co-founder of two spin-off companies: EDALab, focused on networked embedded systems design; and the automation control software company FACTORYAL. 
\end{IEEEbiography}

\begin{IEEEbiography}[{\includegraphics[width=1in,height=1.25in,clip,keepaspectratio]{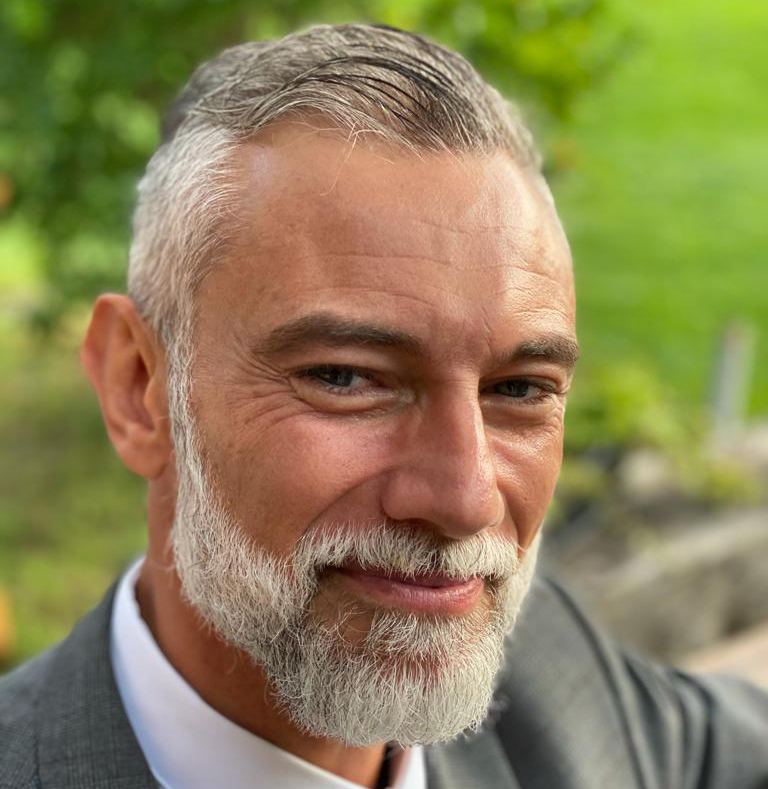}}]{Marco Cristani} (Member, IEEE)
is a Full Professor at the Department of Engineering for Innovation Medicine, University of Verona, Associate Member at the National Research Council (CNR), and External Collaborator at the Italian Institute of Technology (IIT). 
His main research interests are in statistical pattern recognition and computer vision, mainly in deep learning and generative modeling, with application to social signal processing and fashion modeling. 
On these topics, he has published more than 200 papers. 
He has organized 11 international workshops, and cofounded a spin-off company, Humatics, dealing with e-commerce for fashion. 
He is or has been the Principal Investigator of several national and international projects, including PRIN and H2020 projects. 
He is an IAPR fellow. 
\end{IEEEbiography}

\EOD

\end{document}

%% file: srcs/1_intro.tex
\section{Introduction}
\label{cha:1_intro}

A prodigious amount of research has been invested over the past decades in improving embedded technologies in order to enable the use of real-time solutions for many complex and safety-critical applications~\cite{dutta2021tinyml}. 
In this regard, hardware-specific (\eg{}, Edge TPUs) and Micro-Controller Unit (MCU)-based embedded systems have earned a lot of attention, primarily due to their low power requirements, high performance, and, secondarily, for their maintainability, adaptability, and reliability~\cite{capra2019edge}. 
Their integration with sensors enables the perception of the external world, their connection with activators allows different kinds of interventions, and their interconnection unlocks distributed intelligence. 

Embedded technologies are essentially the pillars of the Internet of Things (IoT)~\cite{madakam2015internet} and the associated \textit{smart-X} applications: \textit{smart buildings}~\cite{daissaoui2020iot} and \textit{cities}~\cite{arasteh2016iot}, \textit{smart metering}~\cite{kabalci2016survey}, \textit{agriculture}~\cite{sinha2022recent} and \textit{environment}~\cite{gondchawar2016iot}, \textit{smart health}~\cite{alshehri2020comprehensive}, \textit{smart logistics}~\cite{song2020applications}, and \textit{smart retail}~\cite{jayaram2017smart}. 
More recent advances in the Industrial Internet of Things (IIoT)~\cite{liao2018industrial} have facilitated the real-time intelligent processing of massive amounts of data, promoting fields such as \textit{autonomous driving}~\cite{yurtsever2020survey}, \textit{smart factories}~\cite{hozdic2015smart}, \textit{anomaly detection}~\cite{chandola2009anomaly}, and \textit{predictive maintenance}~\cite{zhang2019data}. 

When we talk about the intelligence of onboard embedded technologies, we mean the learning algorithms that allow devices to make reasoned decisions based on acquired data. 
Unfortunately, Machine Learning (ML) on tiny devices is substantially hard, due to severe architectural, energetic, and latency constraints~\cite{branco2019machine}: the available memory averages a few kilobytes, the accessible power is in the order of milliwatts, and often real-time responses must be guaranteed~\cite{lin2020mcunet}, as in safety-critical systems like health care devices, autonomous driving, or human-robot collaboration in industrial environments, where delayed decisions may have disastrous consequences, ranging from compromised patient well-being and increased road safety hazards, to operational disruptions. 

\begin{figure*}[t!]
    \centering
    \includegraphics[width=.8\linewidth]{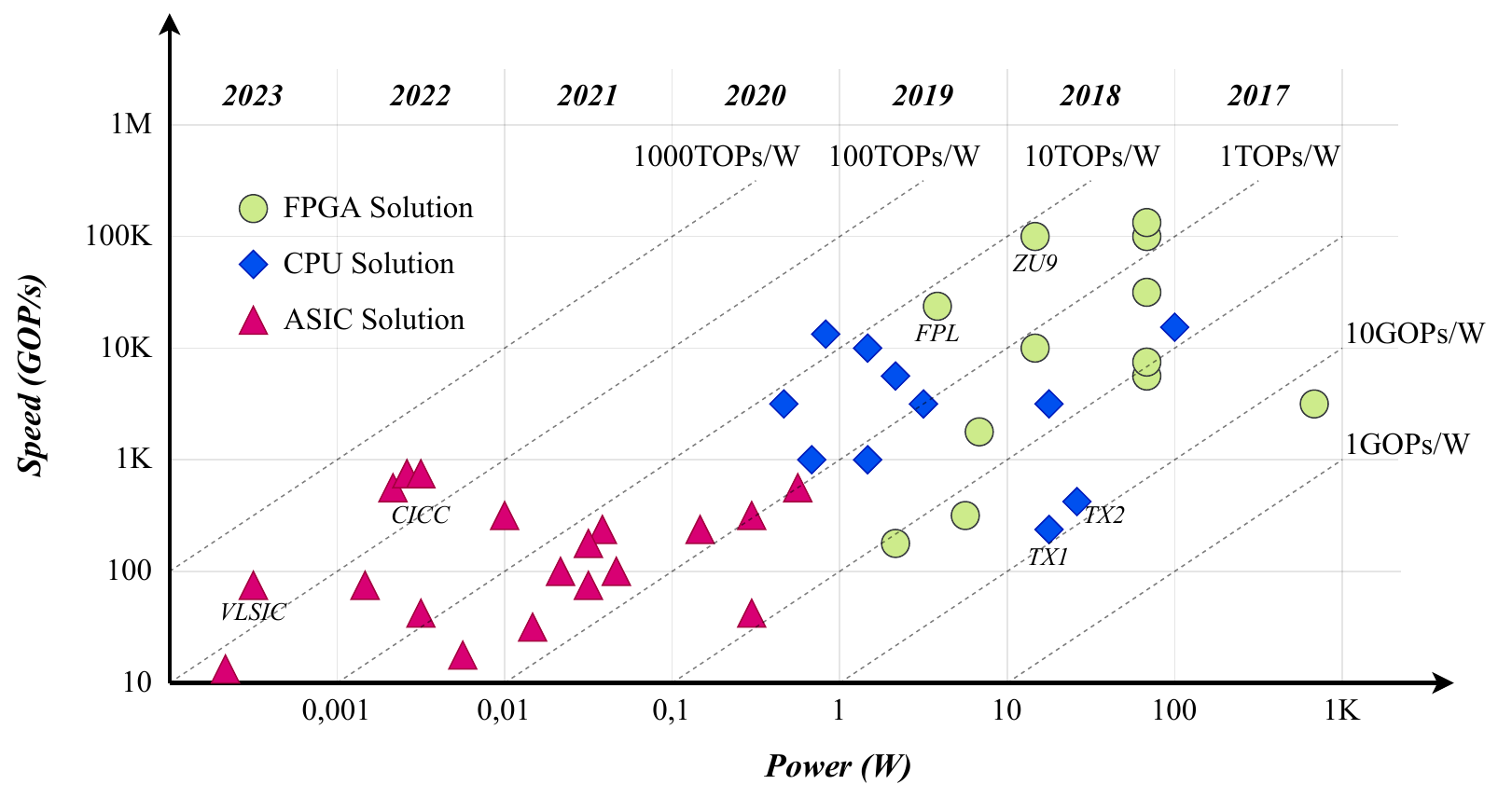}
    \caption{A glance at the latest hardware developed for \tinyml{} reveals a notable trend: recent advances are focused on the goal of minimizing power consumption. 
    This means that the main focus right now is making sure that ML can run on devices with limited resources.}
    \label{fig:figure_1}
\end{figure*}
From these premises, since 2018, the notion of \textit{\textbf{Tiny Machine Learning (\tinyml{})}} has begun to take shape with the following acknowledged definition: \textit{\tinyml{} is a paradigm that facilitates running machine learning at the edge devices with minimal processor and memory requirements; hence, the power consumption of such systems is expected to be within a few milliwatts or less}~\cite{warden2019tinyml}. 
The challenges for \tinyml{} practitioners are formidable: \eg{}, in modern neural networks, among the best currently available technologies, the number of required parameters have skyrocketed to the order of billions~\cite{pramod2021machine}, with larger networks having better results and wider applicability. 
Unfortunately, the energy required to run these networks is proportional to their size, making this trend of scaling up neural networks energetically unsustainable at large scales~\cite{thompson2020computational}: another reason why \tinyml{} has to be considered as a necessary, other than promising, research direction. 
Recent market trends (see Figure~\ref{fig:figure_1}) confirm this rationale: priority has been given to deploying hardware that is less power-hungry, and constraining, therefore, the complexity of implemented learning algorithms: truly, \tinyml{} has to be tinier. 

When it comes to developing a \tinyml{} solution, there are two main classical workflows, namely \textit{\textbf{ML-oriented}} and \textit{\textbf{HW-oriented}}, and a third more recent approach, \textit{\textbf{co-design}}.  
The classical workflows are widely adopted and separate the ML framework design from its hardware incarnation~\cite{dutta2021tinyml,rajapakse2023intelligence}. 
In the first approach, ML experts create, train, and test a suitable model for the problem domain, optimize its parameters, and then deploy this solution on a satisfactory device. 
In the second one, the hardware platform is not prearranged, and development aims to produce optimized hardware by employing specially reduced models and techniques. 

The novel workflow is named co-design because ML experts and hardware engineers are involved together from the start in the design of the solution and actively exchange operational knowledge~\cite{bringmann2021automated}. 
Hardware engineers approach the mathematical notions underlying the ML algorithms and propose suitable hardware components for efficient translations. 
ML researchers meanwhile examine the cutting-edge resources they can exploit, and potentially re-design their algorithms to provide a seamless integration of hardware and software, where the form and the content are malleable and shape each other. 
Here is where shape and content are mixed together, with a blended recipe that constitutes the state-of-the-art of contemporary \tinyml{}. Specifically, in Section~\ref{cha:4_tinyML_pipelines}, we will detail each one of these workflows.

\subsection{Motivation and Contributions}
\begin{figure*}[t!]
    \centering
    \includegraphics[width=.9\linewidth]{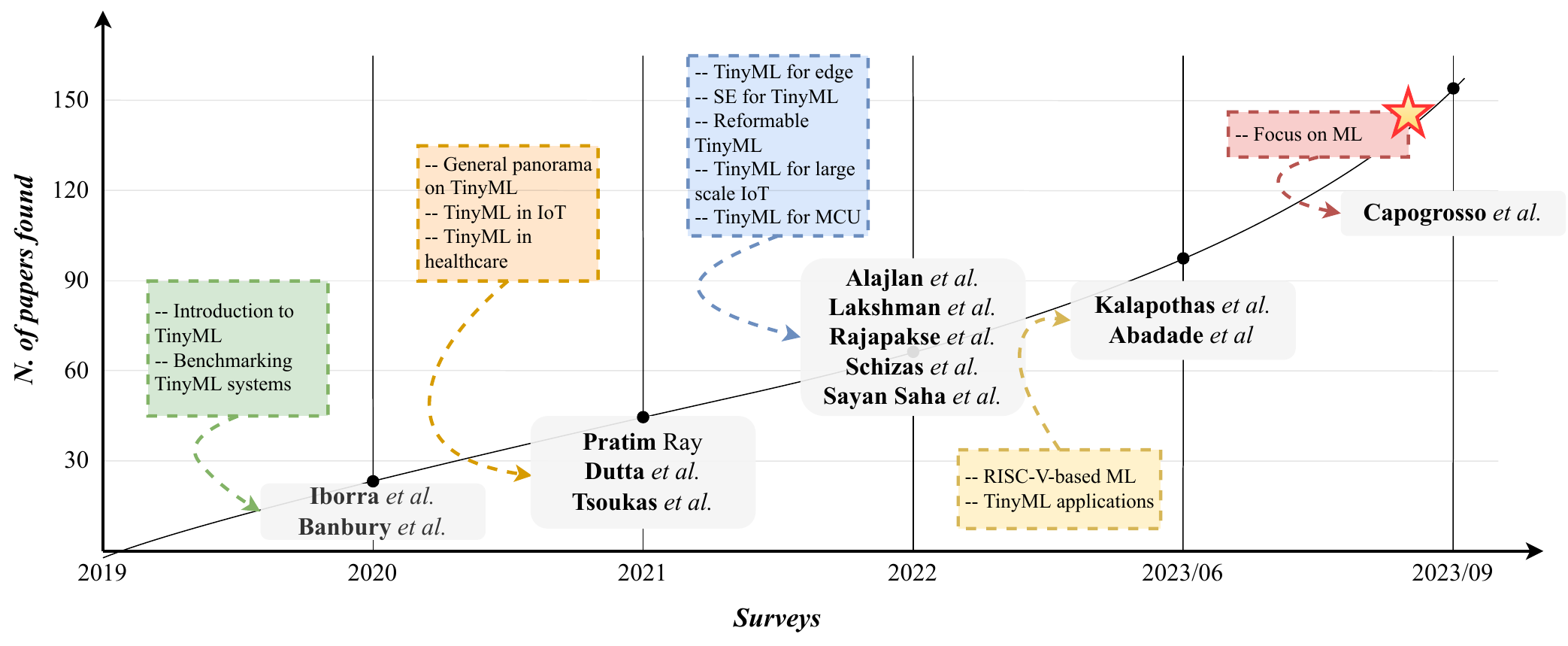}
    \caption{The number of papers on \tinyml{} published so far, and the surveys on the topic.
    As evident, there is an exponential growth in the number of research papers, and it's worth noting that our survey not only stands as the most recent but also uniquely concentrates on the ML perspective, distinguishing it from all other existing surveys in the field.}
    \label{fig:figure_2}
\end{figure*}
We offer two contributions with this survey: we first provide an up-to-date overview of the rapidly evolving state-of-the-art in the field of \tinyml{}. 
Since the number of research articles published on \tinyml{} is increasing exponentially (see Figure~\ref{fig:figure_2}), the number of surveys and papers on the subject is following suit. 
Specifically, we catalog the literature up to September 2023. 

As a further and unique contribution, this survey emphasizes the ML point of view, not only reporting the very latest in \tinyml{} frameworks but also suggesting recent variations and advancements in the ML technologies that a \tinyml{} practitioner may want to explore to improve on the state-of-the-art: \eg{}, topics like meta-learning~\cite{hospedales2021meta}, Rational Activation Functions (RAFs)~\cite{trimmel2022era}, and Versatile Learned Optimizers (VeLO)~\cite{metz2022velo}. 
In this regard, we provide some insights into these ML methodologies and we additionally address the most recent developments as potential \tinyml{} future breakthroughs.

\subsection{Article Organization}
The survey is organized as follows.  
Section~\ref{cha:2_existing_surveys} provides an extensive overview of the existing surveys on \tinyml{}, indicating the differences with this work, while Section~\ref{cha:3_selection_criteria} describes the article selection criteria used in creating this systematic review. 
Section~\ref{cha:4_tinyML_pipelines} clarifies what it means to design a \tinyml{} solution in terms of workflow, as we briefly explained in the previous subsection. 
Section~\ref{cha:5_efficient_dl} is the core of the survey and presents the collection of algorithms and techniques to enable efficient ML on tiny devices. 
Section~\ref{cha:6_devices_and_tools} reports several hardware specifications, libraries, and software platforms that are state-of-the-art for \tinyml{} application development. 
Section~\ref{cha:7_discussion} summarizes the overview and proposes potential future directions. 
Finally, Section~\ref{cha:8_conlusion} concludes the paper. 

%% file: srcs/2_existing_surveys.tex
\section{Existing Surveys}
\label{cha:2_existing_surveys}

In this section, we provide discussions with the most recent surveys~\cite{banbury2020benchmarking,sanchez2020tinyml,siang2021anomaly,tsoukas2021review,dutta2021tinyml,ray2022review,njor2022primer,giordano2022survey,immonen2022tiny,schizas2022tinyml,lakshman2022software,alajlan2022tinyml,bamoumen2022tinyml,su2022ai,saha2022machine,rajapakse2023intelligence,prakash2023cfu,kalapothas2023survey,abadade2023comprehensive}.

\subsection{Hardware Perspective}
From the point of view of hardware technologies, we can distinguish \tinyml{} solutions based on \textit{\textbf{Application-Specific Integrated Circuits (ASICs)}}, \textit{\textbf{(MCUs)}}, and \textit{\textbf{Field Programmable Gate Arrays (FPGAs)}}, in increasing order of power consumption (see Figure~\ref{fig:figure_1}). 
Specific surveys exist for each one of these technologies. 

\textit{ASIC} is the focus of~\cite{kalapothas2023survey}. 
The RISC-V, \ie{}, the fifth generation of the Berkeley Reduced Instruction Set Computer (RISC) architecture, has been widely adopted by many researchers and commercial users, with several openly available implementations to choose from. 
Selecting the appropriate combination of RISC-V processor cores, architectures, configurations, and ML software frameworks is not trivial. 
In order to facilitate this process, the survey discusses the various RISC-V-based hardware implementations, in terms of available cores and System-On-Chip (SoC), in conjunction with the software frameworks and software stacks for the SoC generation. 
It includes a review of the latest released frameworks supporting open hardware integration for ML applications. 

\textit{MCU} and \tinyml{} are the subjects of~\cite{sanchez2020tinyml,giordano2022survey,immonen2022tiny}. 
In~\cite{sanchez2020tinyml}, the authors analyze the \tinyml{} frameworks for integrating ML algorithms within MCUs and present a real-world case study. 
They first give a small overview of the ecosystem of applications in which \tinyml{} techniques can be applied and then highlight the opportunities in various sectors currently undergoing a digital transformation. 
Finally, they propose a Multi-Radio Access Network (Multi-RAT) architecture for smart frugal objects: \ie{}, sporadically messaging interconnected devices with constrained resources. 

In~\cite{giordano2022survey}, the authors focus on surveying, comparing, and evaluating seven different recent and popular MCUs on a face recognition task based on a Convolutional Neural Network (CNN) workload. 
Their evaluation considers four key metrics (\ie{}, power efficiency, energy per inference, inference efficiency, and inference time) that can be used to benchmark ML applications on MCU-based devices. 

\textit{FPGA} is considered in~\cite{prakash2023cfu}. 
The authors present ``CFU Playground'', a full-stack open-source framework that enables the rapid and iterative design of ML accelerators for embedded ML systems through custom function units (CFU), \ie{}, hardware that augments the standard functions of a CPU. 
This toolchain integrates open-source software, register transfer level generators, and FPGA tools for synthesis, place, and route. 
To illustrate their approach, they apply their methodology to two common \tinyml{} use cases: image classification and keyword spotting. 
In the first case, they show how to obtain iterative hardware-software improvements with ease, and, in the second, how to co-optimize the CPU and the CFU together in severely resource-constrained environments.

\subsection{Application Perspective}
In relation to applied \tinyml{}, we review surveys in the fields of \textit{\textbf{IoT}}, \textit{\textbf{environmental challenges}}, \textit{\textbf{predictive maintenance (PdM)}}, \textit{\textbf{anomaly detection}}, and \textit{\textbf{healthcare}}. 

In~\cite{dutta2021tinyml}, the authors provide background information on the benefits that \tinyml{} can offer to the \textit{IoT} panorama, such as low latency, effective bandwidth utilization, strengthened data safety, and enhanced privacy. 
Then, they show how to implement \tinyml{}-as-a-service, \ie{}, an IoT device that concretely takes part in the execution of intelligent services. 
In~\cite{schizas2022tinyml}, the authors explore the integration of \tinyml{} with network technologies such as 5G and LPWAN. Ultimately, we anticipate that this analysis will serve as an informational pillar for the IoT/cloud research community and pave the way for future studies. 

Of particular interest in recent years, \tinyml{} has been applied to \textit{environmental challenges}~\cite{bamoumen2022tinyml} such as global warming, climate change, natural resource scarcity, and pollution monitoring. 
With their ability to deploy intelligent analysis together with sensing devices, \tinyml{} provides the natural evolution to data gathering in the environmental domain to protect our societies and the natural world. 
This survey elaborates on the role of \tinyml{} devices and their limit in this context. 

In~\cite{njor2022primer}, the authors investigate techniques used to optimize \tinyml{}-based \textit{PdM} systems. 
They describe PdM, and how \tinyml{} can provide an alternative to cloud-based PdM, showing commonly used libraries, hardware, datasets, and models. 
Furthermore, they show known techniques for optimizing \tinyml{} models. 

\textit{Anomaly detection} is the detection of unexpected patterns in the data. 
In~\cite{siang2021anomaly}, the authors highlight the state-of-the-art current works on \tinyml{} for anomaly detection, providing suggestions on the research direction, and introducing potential future endeavors. 

An essay on \tinyml{} approaches for \textit{healthcare} is presented in~\cite{tsoukas2021review}. 
The authors collect references related to \textit{i)} the selection of patients for investigation, monitoring, and protocol adherence, \textit{ii)} the collection, processing, analysis, and management of data, and \textit{iii)} drug validation trials, followed by the solutions they bring, especially using wearable devices. 

Finally, in~\cite{abadade2023comprehensive}, the authors present an overview of many \tinyml{} \textit{applications} and related research efforts. 
Specifically, the survey builds a taxonomy of \tinyml{} techniques that have been used so far to bring new solutions to various domains, such as healthcare, smart farming, environment, and anomaly detection.

\subsection{Field Viewpoint}
In~\cite{banbury2020benchmarking}, the authors discuss the challenges and directions toward developing a fair and useful \tinyml{} \textit{benchmarking} suite. 
The group has selected four use cases to target: audio wake words, visual wake words, image classification, and anomaly detection. 
For each of the use cases, reference datasets, and baselines were also selected. 
The benchmarking suite provides results in terms of the accuracy of the model, inference latency, and energy consumption. 
Notably, this is one of the few literature reviews that presents datasets that can be particularly useful for the benchmarking and design of \tinyml{} systems. 

In~\cite{ray2022review} and~\cite{alajlan2022tinyml}, the authors present the \textit{background} of \tinyml{}, list the tool for supporting \tinyml{}, and the key enablers (\eg{}, model compression and quantization) for the improvement of \tinyml{} systems. 
However, neither of them is focused on advanced learning aspects for \tinyml{}, as well as established HW-SW co-design support to enhance \tinyml{} systems, making our survey clearly distinct from these two. 

In~\cite{lakshman2022software}, the authors aggregate the key challenges reported by \tinyml{} developers and identify state-of-art \textit{Software Engineering (SE)} approaches that can help address key challenges in \tinyml{}-based IoT embedded vision. 
Examples of these challenges include the lack of curated datasets derived from IoT-embedded vision sensors, the application portability across different devices and different vendors, and the compiler choices, since embracing sophisticated compilers can help optimize for specific MCU targets. 
However, this affects portability, and hence, it challenges large-scale deployment under availability constraints. 

In~\cite{su2022ai} the authors focus on edge \textit{training} and edge \textit{inference}. 
This paper provides a survey of existing architectures, technologies, frameworks, and implementations in these two areas, and discusses existing challenges, possible solutions, and future directions. 

Finally, in~\cite{saha2022machine}, a review of \textit{deployment techniques} for \tinyml{} devices is provided, with also numerical insights to prove which deployment workflow is more promising given the constraints of the models and input data (\eg{}, sparsity, compression, \etc{}). 
They inspect the engineering of reducing the computation and memory footprint for the inference of already existing models. 
Furthermore, they set up some case studies to present the deployment of several famous models with and without various techniques (such as compression and feature projection), with numerical results to highlight the benefits of each technique. 

Most of the works previously listed assume that \tinyml{} can only run inference on data. 
Despite this, growing interest in \tinyml{} has led to work that makes them \textit{reformable}, \ie{}, work that permits \tinyml{} to learn from new data points once deployed. 
This originates from the need to combat model drift – the inevitable degradation of a model’s performance due to the ever-changing nature of data. 
In~\cite{rajapakse2023intelligence}, the authors provide a survey on reformable \tinyml{} solutions. 

It should be noted that the majority of these works are not conducted based on a well-defined and widely known Systematic Literature Review (SLR), \eg{}, using Preferred Reporting Items for Systematic reviews and Meta-Analyses (PRISMA)~\cite{yepes2021prisma}, except~\cite{lakshman2022software} and \cite{abadade2023comprehensive}. 

%% file: srcs/3_selection_criteria.tex
\section{Selection Criteria}
\label{cha:3_selection_criteria}

\begin{figure}[t!]
    \centering
    \includegraphics[width=.9\linewidth]{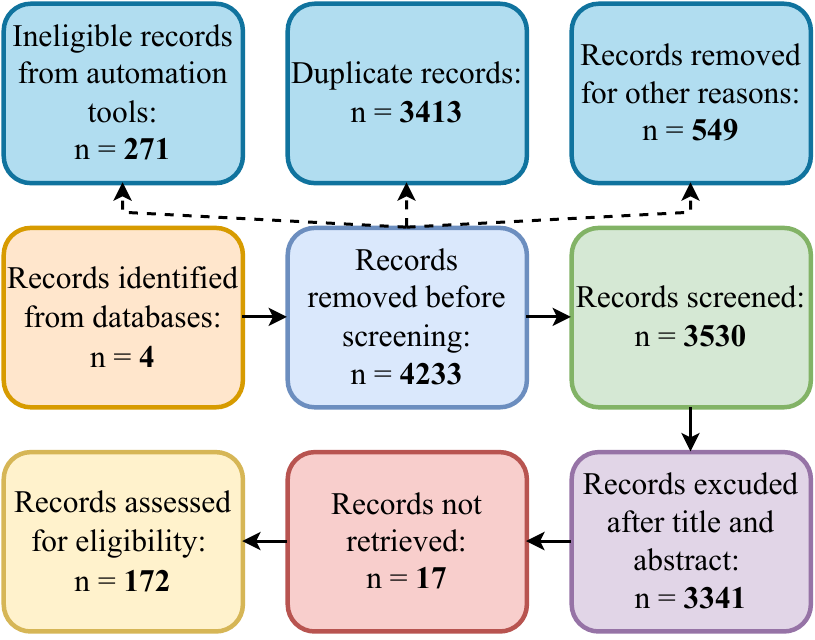}
    \caption{\textsc{PRISMA}-based flowchart of the retrieval process. The image has been changed from the standard flowchart solely for aesthetic purposes.}
    \label{fig:PRISMA-on-tinyML}
\end{figure}

This section describes the selection criteria of this systematic review, \ie{}, how the papers that were considered were selected. 

Only publications in the English language were considered, and all studies had to be published in peer-reviewed journals or conference proceedings (except for the \tinyml{} software tools presented in Section~\ref{sec:6_e_sw_tools}). 
The search strategy and selection criteria were developed in consultation with all authors. 
Any disagreements between authors were resolved through discussion and consensus. 
To gather up-to-date knowledge from a broad spectrum of information sources, this comprehensive ML-oriented survey on \tinyml{} was conducted following a widely known SLR methodology based on the PRISMA guidelines, the golden standard for improving transparency, accuracy, and completeness in documented systematic reviews and meta-analyses. 

The included studies were extracted from the following five databases: \textit{\textbf{Web of Science}}, \textit{\textbf{Scopus}}, \textit{\textbf{IEEE Xplore}}, \textit{\textbf{ScienceDirect}}, and \textit{\textbf{Google Scholar}}, from January 2018 to September 2023. 
All searches included the following terms: \textit{``\tinyml{}''}, \textit{``efficient machine deep learning''}, \textit{``neural network optimization''}, \textit{``iot machine deep learning''}, \textit{``embedded machine deep learning''}, \textit{``edge machine deep learning''} and \textit{``mcu machine deep learning''}. Therefore, all the cited papers in this work were found using the above keyword combination. 

Our keywords produced a total of 4233 records. 
Figure~\ref{fig:PRISMA-on-tinyML} illustrates the PRISMA flowchart, which serves as a transparent and replicable means of reporting the systematic review’s search and selection process. 
First, we removed all duplicate papers (3413 excluded). 
Next, we excluded all the papers marked as ineligible by the automation tool (271 excluded), and not accessible papers (\eg{}, requiring paid access) (549 excluded). 
After the title and abstract screening process, a total of 189 articles were selected (3341 excluded). 
The number of records not found is 17. 
As a result, 172 were eligible. 

Finally, out of the 172 reviewed papers, none of them were found to be survey papers on ML-oriented techniques for \tinyml{}. 
As a result, we claim that this is the first systematic review to address this topic. 

%% file: srcs/4_tinyML_pipelines.tex
\section{TinyML Workflows}
\label{cha:4_tinyML_pipelines}

In this section, we present an overview of how \tinyml{}-based systems are built. 
The two intrinsic ingredients of such systems are the ML model and the hardware platform, therefore the natural approach for developers in the field is to start working from the most familiar component. 
A more efficient, but challenging, alternative is to develop both sides from the beginning and create an integrated solution. 
As anticipated in Section~\ref{cha:1_intro}, the two traditional workflows for \tinyml{} solutions are \textit{\textbf{ML-oriented}} and \textit{\textbf{HW-oriented}}, while the holistic methodology is called \textit{\textbf{co-design}}. 

\begin{figure*}[t!]
    \centering
    \includegraphics[width=0.8\linewidth]{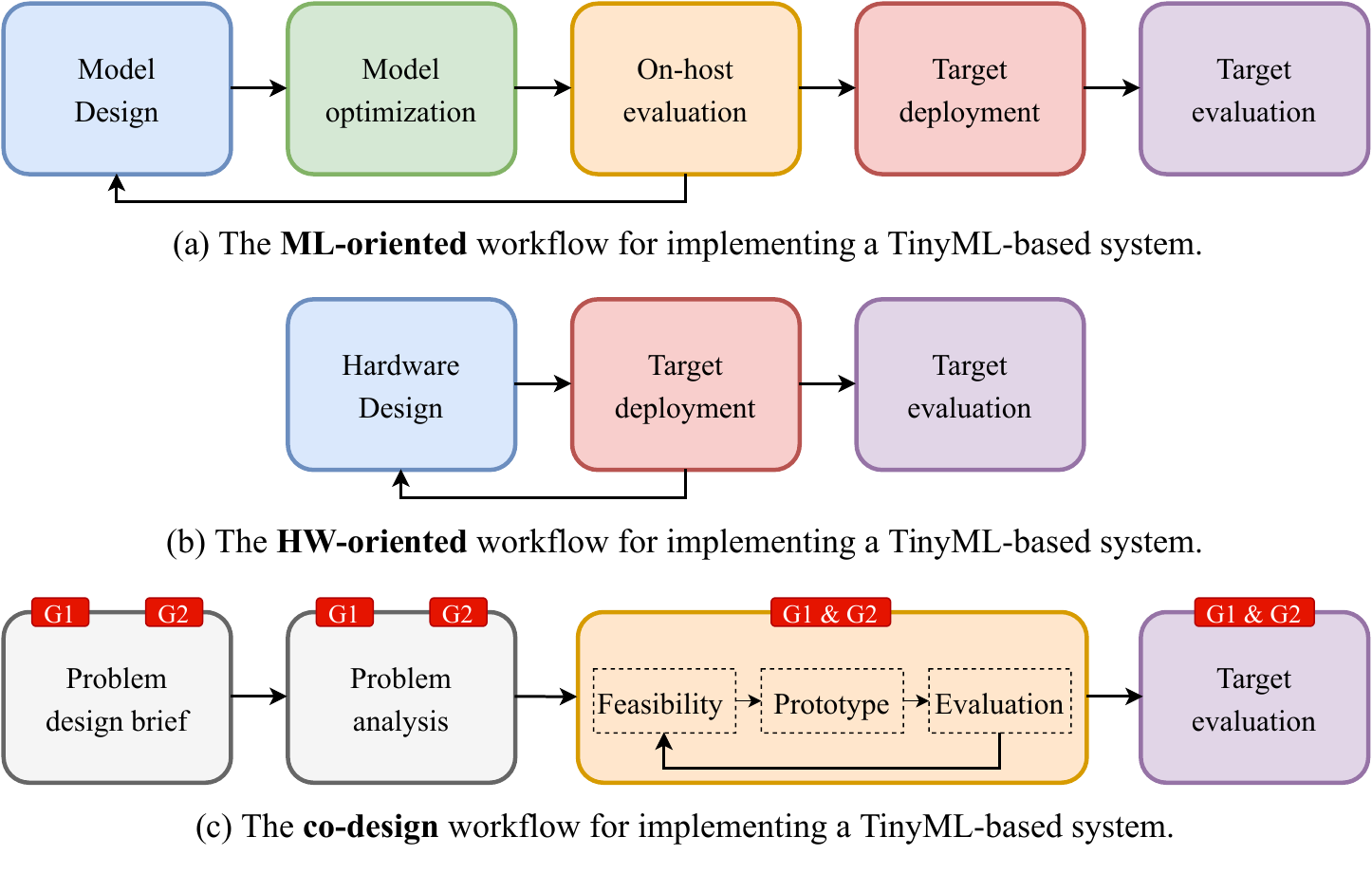}
    \caption{The three workflows for implementing a \tinyml{}-based system: (a) ML-oriented, (b) HW-oriented, and (c) co-design. 
    In the latter, two separate working groups G1 (ML specialists) and G2 (HW specialists) collaborate together. 
    }
    \label{fig:workflows}
\end{figure*}

In the \textit{ML-oriented} workflow (see Figure~\ref{fig:workflows}.a), the majority of the expertise is in the design, adaptation, training, and evaluation of ML models, while the choice of hardware platforms is fixed or limited, due to necessity or specific industrial requirements~\cite{iccad2022tinymlchallenge,tinyml2022tinymlchallenge,ren2023tinyreptile}. 
A typical example of this workflow is the porting of modern neural network models to embedded devices~\cite{heim2021measuring}. 
This requires extensive experimental investigations for the implementation to be efficient in terms of power consumption, latency, and memory usage, all resources in short supply on such devices compared to cloud solutions. 

In particular, we identify the following stages in the ML-oriented workflow: 
\begin{itemize}
\item \textbf{Model design}: ML practitioners formulate, train, and validate a comprehensive model suitable for the problem domain. 
This stage is highly dependent on the nature of this domain but disregards on purpose the specifics of the hardware platform to achieve maximum generalization and performance. 
\item \textbf{Model optimization}: This stage consists of different strategies to compromise performance for efficiency, discussed in more detail in Section~\ref{cha:5_efficient_dl}. 
\item \textbf{On-host evaluation}: The optimized model is evaluated against the performance parameters required in the specifications and, if found lacking, it is re-designed. 
\item \textbf{Target deployment}: Specialized optimizations are applied to the model in order to increase the inference efficiency by leveraging specific features of the hardware device. 
\item \textbf{Target evaluation}: The final evaluation of the system in production is performed. 
\end{itemize}

On the other hand, in the \textit{HW-oriented} approach (Figure~\ref{fig:workflows}.b), the developers are mainly focusing on designing enhanced hardware platforms that are optimized for embedded applications, in order to run current and future state-of-the-art ML algorithms. 
This often involves investigating the bottlenecks in an existing architecture with regard to computations within a ML framework, like neural networks, and the design of hardware accelerator modules to improve throughput and consumption: \eg{}, reducing computational complexity in convolution layers~\cite{chang2018reducing,olyaiy2021accelerating}, efficient, low-power and feature-rich perceptrons~\cite{lin2021efficient}, enhanced data caches~\cite{zhou2020enhanced}. 
In other cases, the developers design new hardware platforms optimized for embedded applications with extended digital signal processing capabilities already integrated~\cite{conti2014energy}. 
These in turn require the development of optimized computing libraries~\cite{garofalo2020pulp,garofalo2019pulp} to extract the most performances. 

A HW-oriented workflow may have the following stages:
\begin{itemize}
\item \textbf{Hardware design}: Hardware practitioners create the design for an architecture, or accelerator module in an architecture, that improves performances for a given class of computing problems or signal processing algorithms. 
\item \textbf{Target deployment}: Assessment of the performance of the optimized hardware on benchmarks of the given computing problems, mostly in simulated environments. 
In case of unsatisfying results, return to the design stage. 
\item \textbf{Target evaluation}: Production and evaluation of the physical hardware devices. 
\end{itemize}

Finally, in the \textit{co-design} workflow (Figure~\ref{fig:workflows}.c), the approach is to integrate both sides of the development from the start in order to gain further improvements in performance and resource consumption. 
In particular, while model optimization and hardware design are separate steps in the previous workflows (Figure~\ref{fig:workflows}.a and Figure~\ref{fig:workflows}.b), here they are intertwined and co-optimized: in some cases to create bespoke architectures for specific ML algorithms on FPGAs~\cite{prakash2023cfu}, in other cases to allow neural network computations on customized accelerators using analog compute-in-memory (CiM) hardware through HW-informed training methodologies~\cite{zhou2022ml}. 

The co-design workflow may be described with the following steps: 
\begin{itemize}
\item \textbf{Problem design brief}: Two separate working groups G1 (ML specialist) and G2 (HW specialist) define the capabilities and requirements of the target device. 
\item \textbf{Problem analysis}: The two groups specify their state-of-the-art architecture after exploring the possible alternatives. 
\item \textbf{Co-design step}: In a cooperative and concurrent design process, specific hardware and software components for selected sections of an application must be chosen with a global view of the system. 
\item \textbf{Target evaluation}: Final evaluation of the model-specific and target-specific optimizations for the device in production. 
\end{itemize}

%% file: srcs/5_efficient_dl.tex
\section{Learning panorama under the TinyML lens}
\label{cha:5_efficient_dl}

\begin{figure*}[t!]
    \centering
    \includegraphics[width=0.8\linewidth]{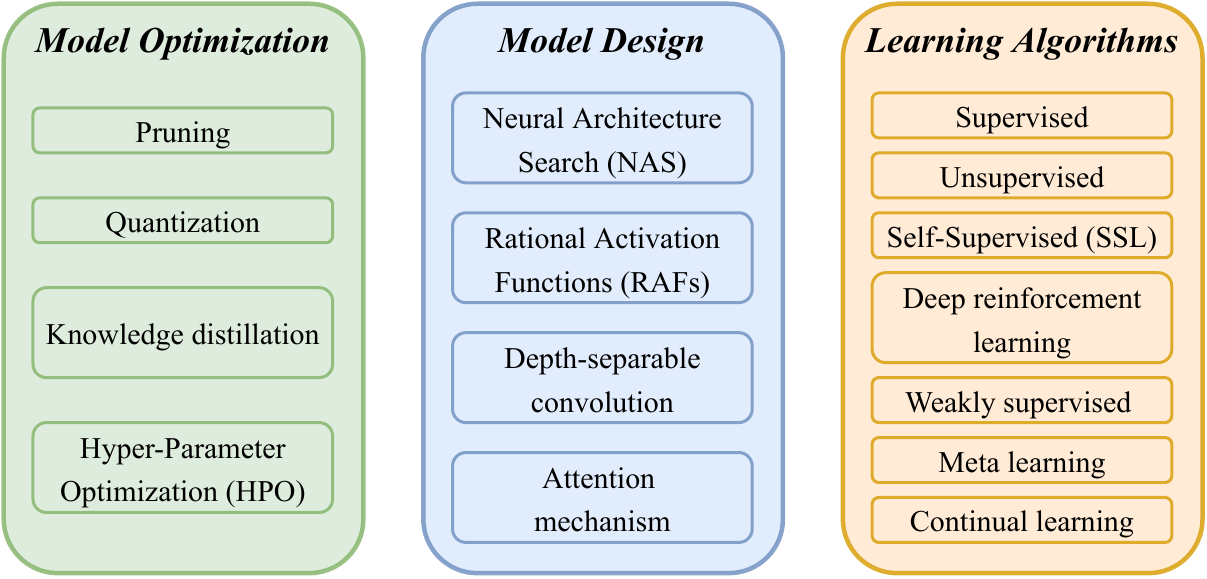}
    \caption{Our proposed taxonomy, which covers the learning panorama under the \tinyml{} lens, includes three key domains: model optimization, model design, and learning algorithms.}
    \label{fig:efficient_DL}
\end{figure*}

To leverage the full potential of \tinyml{}, exploring and understanding the complexities involved in designing and optimizing ML models for specifically resource-limited devices is essential. 
We propose the taxonomy shown in Figure~\ref{fig:efficient_DL}, which covers the learning panorama under the \tinyml{} lens. 
In the following sections, we will delve into the macro-areas of \textit{\textbf{model optimization}}, \textit{\textbf{model design}}, and \textit{\textbf{learning algorithms}}.

\subsection{Model Optimization}
\begin{table}[t!]
    \centering
    \caption{Popularity and their respective papers over the years of model optimization techniques based on referenced research contributions.}
    \begin{scriptsize}
    \input{tables/t_model_optimization}
    \end{scriptsize}
    \label{tab:t_model_optimization}
\end{table}
Model optimization techniques tailored for \tinyml{} generally produce smaller memory footprints, lower energy consumption, and reduced inference latency. 
In many cases, \eg{}, large neural network models, there are simply not enough resources available on embedded devices. 
The following paragraphs explore techniques such as \textit{\textbf{pruning}}, \textit{\textbf{quantization}}, \textit{\textbf{knowledge distillation}}, and \textit{\textbf{Hyper-Parameter Optimization (HPO)}}, that enable efficient model deployment. 
Table~\ref{tab:t_model_optimization} provides an overview of the popularity of these techniques within the \tinyml{} field, considering the referenced research contributions. 

\subsubsection{Pruning}
This process eliminates weight connections within a network to accomplish different goals like reduced model footprint and accelerated inference speed~\cite{vadera2022methods}. 
Despite a lack of standardized benchmarks and metrics, due to differences in goals favoring different design choices and evaluations, pruning is effective at compressing models while keeping (or sometimes increasing) accuracy~\cite{blalock2020state}. 
Pruning techniques can be applied during the training process or after the model has been trained. 
During training, pruning regularizes the model, mitigating the risk of overfitting~\cite{zhu2017prune}. 
Post-training pruning is instead employed to eliminate redundant connections and parameters from the model, thereby enhancing its efficiency and accelerating its execution~\cite{liu2018rethinking}. 
We identified in the current literature three main approaches: \textit{\textbf{weight pruning}}, \textit{\textbf{neuron pruning}}, and \textit{\textbf{structured pruning}}. 

Specifically, \textit{weight pruning} is a technique that eliminates connections or weights in a model falling below a given threshold for weight size~\cite{yu2017scalpel}. 
Approaches based on this technique are gaining interest due to their immediate applicability~\cite{sun2023case,hashir2023tinyml}. 
Similarly, \textit{neuron pruning} discards entire neurons based on a given threshold of importance~\cite{hu2016network} and \textit{structured pruning} removes entire structures or sub-networks from a model~\cite{anwar2017structured}. 

\subsubsection{Quantization}
This involves performing computations and storing tensors at lower bit widths compared to floating point precision~\cite{gholami2022survey}. 
By utilizing fewer bits to represent data, such as 16-bit floats or 8-bit integers instead of 32-bit floating-point numbers, quantization enables more compact model representations and the utilization of efficient vectorized operations on various hardware platforms~\cite{krishnamoorthi2018quantizing}. 
This technique is particularly beneficial during inference, significantly reducing computation costs while maintaining inference accuracy~\cite{cai2017deep}.   
Quantization can be achieved through two approaches: \textit{\textbf{Quantization-Aware Training (QAT)}}, which involves re-training the model, and \textit{\textbf{Post-Training Quantization (PTQ)}}, which applies quantization without re-training. 

\textit{QAT} involves quantizing a pre-trained model and subsequently performing a fine-tuning step to recover any accuracy loss caused by quantization-related errors, which may impact model performance~\cite{nagel2022overcoming}. 
The QAT process consists of two stages: pre-training and fine-tuning. 
In the pre-training stage, the network is trained using standard techniques in a full-precision floating-point format (32-bit) to learn data patterns and develop robust feature representations. 
In the fine-tuning stage, the network is converted to a quantized representation, combining fixed-point and floating-point arithmetic. 
This adjustment allows the network to adapt to the quantized representation while preserving accuracy. 
QAT encompasses different methods, including \textit{hybrid}~\cite{mishchenko2019low}, \textit{layer-wise}~\cite{tang2022mixed}, and \textit{adaptive} approaches~\cite{youn2022bitwidth}. 

\textit{PTQ} reduces memory usage and computational costs by converting model weights and activations from high-precision floating-point to low-precision numbers~\cite{jacob2018quantization}. 
Initially, the model is trained using floating-point representation, followed by quantization of weights and activations using techniques like k-means clustering or vector quantization~\cite{nagel2020up}. 
The adoption of low-precision numbers, such as 8-bit integers, significantly reduces memory requirements, enabling more efficient model execution and suitability for resource-constrained environments~\cite{wang2020towards}. 
These techniques are also known as \textit{Dynamic Range Quantization (DRQ)} or \textit{Full-Integer Quantization (FIQ)}, depending on whether only the weights are being quantized to 8-bit integers or inputs and activations functions. 

Like in pruning, the application of quantization techniques allows for immediate deployment of already existing models to resource-constrained devices in various fields like computer vision~\cite{moosmann2023tinyissimoyolo,lu2023enhancing} and healthcare~\cite{alajlan2023ddd}. 

\subsubsection{Knowledge Distillation} \label{sub:kd}
This technique transfers knowledge from a large, complex model (teacher) to a smaller, simpler model (student)~\cite{gou2021knowledge}. 
This process is important for various reasons, such as reducing computational demands or enhancing model performance on specific tasks. 
Knowledge types, distillation strategies, and teacher-student architectures are vital factors in student learning during knowledge distillation. 
The subsequent paragraphs introduce the key categories of \textit{\textbf{knowledge types}} and \textit{\textbf{distillation strategies}}. 

The extraction of \textit{knowledge} from teachers and its utilization for training student networks can be classified into three categories: \textit{\textbf{response-based}}, \textit{\textbf{feature-based}}, and \textit{\textbf{relation-based}}. 
Specifically, \textit{response-based} knowledge distillation involves mimicking the final predictions of the teacher model by capturing the neural response in the last output layer~\cite{dai2021general}. 
\textit{Feature-based} knowledge expands upon this approach by using both the outputs of the last layer and intermediate layers to train thinner networks~\cite{zhang2020improve}. 
Finally, \textit{relation-based} knowledge takes a step further by exploring the relationships between different layers or data samples in addition to the outputs of specific layers in the teacher model~\cite{cheng2021relation}. 

The \textit{distillation} schemes are also crucial for the student learning process. 
Depending on the training strategy, the following three different categories are presented: \textit{\textbf{offline distillation}}, \textit{\textbf{online distillation}}, \textit{\textbf{self-distillation}}. 
\textit{Offline distillation} is a two-stage strategy, where the teacher model is first trained on a set of training samples, and then the trained teacher model is used to guide the student model by extracting intermediate features or logits~\cite{zhao2020highlight}. 
On the other hand, \textit{online distillation} is an end-to-end approach where both the teacher and student models are updated simultaneously, making it suitable when the teacher model is not significantly larger or higher performing~\cite{zhang2021adversarial}. 
Finally, \textit{self-distillation} is a special case of online distillation where the teacher and student networks have the same architecture~\cite{yun2020regularizing}. 

In general, knowledge distillation is used to achieve a good trade-off between small model size and an acceptable accuracy~\cite{al2022implementation}. 
For this reason, it is widely adopted in several fields where existing models are well-performing but unable to be deployed ``as they are'' in resource-constrained hardware. 
This is the case with large scaling requirements~\cite{brutti2022optimizing}, bandwidth-limited domains~\cite{korber2021tiny}, and healthcare applications, where the trade-off between accuracy and model size needs to produce a high accuracy model that can fit the hardware requirements~\cite{ukil2021resource}. 

\subsubsection{Hyper-Parameter Optimization (HPO)}
This technique automates the search for the optimal hyper-parameter values of a model to enhance its performance on a specific task~\cite{bergstra2011algorithms}. 
Hyper-parameters, such as learning rate, batch, and network size, are predetermined parameters that influence model behavior~\cite{wu2019hyperparameter}. 

HPO utilizes search algorithms, such as \textit{Grid Search}, \textit{Random Search}, and \textit{Bayesian Optimization}, to explore the hyper-parameter space and identify the combination that yields the best performance~\cite{bergstra2012random}. 
By automating the tuning process, HPO reduces the effort and time required while improving the model's performance.

\subsection{Model Design}
\begin{table}[t!]
    \centering
    \caption{Popularity and their respective papers over the years of model design techniques based on referenced research contributions.}
    \begin{scriptsize}
    \input{tables/t_model_design}
    \end{scriptsize}
    \label{tab:t_model_design}
\end{table}
Unlike traditional ML models, \tinyml{} models require careful design considerations to strike a balance between accuracy and efficiency. 
This section investigates techniques in model architecture exploration, model simplification, and architectural modifications that provide lightweight models, yet are capable of delivering acceptable performances for their intended applications. 
In the following paragraphs we explore \textit{\textbf{Neural Architecture Search (NAS)}}, \textit{\textbf{Rational Activation Functions (RAFs)}}, \textit{\textbf{depth-separable convolution}}, and the \textit{\textbf{attention mechanism}}. 
Table~\ref{tab:t_model_design} provides an overview of the popularity of these techniques within the \tinyml{} field, considering the referenced research contributions. 

\subsubsection{Neural Architecture Search (NAS)} 
Neural architecture design plays a crucial role in data representation and performance, but it heavily relies on researchers' knowledge and experience. 
NAS automates the process of discovering optimal architectures for specific needs, replacing manual tweaking with an automated exploration of more complex architectures. 

NAS utilizes search algorithms, such as \textit{\textbf{reinforcement learning}}, \textit{\textbf{evolutionary algorithms}}, and \textit{\textbf{gradient-based methods}}, to identify architectures that maximize performance on a given task. 
Moreover, in~\cite{njor2023data}, the authors argue that it is beneficial to NAS approaches for resource-constrained systems to also search for appropriate data granularity. 
Specifically, data granularity refers to the concept that data can be fed into an ML model at various levels of detail (\eg{}, an audio sample can be presented to an ML model using different sample rates). 
By automating the search process, NAS reduces the time and effort required for network design and optimization, leading to improved task performance~\cite{ren2021comprehensive,baymurzina2021review,liu2021survey,pau2023quantitative}. 

For example, in these works~\cite{banbury2021micronets,liberis2021munas,garavagno2023hardware}, NAS algorithms targeted specifically to microcontrollers are investigated, demonstrating that NAS promises to help design accurate ML models that meet the tight MCU memory, latency, and energy constraints~\cite{mendis2021intermittent}. 

\subsubsection{Rational Activation Functions (RAFs)}
Activation functions play a central role in deep learning since they form an essential building stone of neural networks, thus, individuating new activation functions that can potentially improve the results is still an open field of research. 
Recently, RAFs have awakened interest because they were shown to perform on par with state-of-the-art activations on image classification~\cite{molina2019pad}. 
They are trainable in an end-to-end fashion using backpropagation and can be seemingly integrated into any neural network in the same way as common activation functions (\eg{}, ReLU). 
In other words, the key idea is to involve the activation functions in the learning process together (or separately) with the other parameters of the network such as weights and biases. 

RAFs have several advantages over standard activation functions~\cite{apicella2021survey}. 
For example, they can provide better approximation capabilities, which can improve the performance of the neural network~\cite{trimmel2022era}. 
Additionally, RAFs can have more flexible shapes, making them better suited for modeling a wider range of data distributions. 

Thus, by exploring RAFs, which can potentially strike a balance between accuracy and computational cost, we could unlock new avenues for creating compact yet high-performing models ideal for resource-constrained contexts. 
Despite the scarcity of previous research on the subject, examining RAFs might lead to ground-breaking findings and innovative insights in refining \tinyml{} models for real-world applications. 

\subsubsection{Convolutional Layers}
In the convolution operation, each filter convolves over the spatial dimensions and channel dimensions of the input. 
The filter size is typically denoted as $s_x\times{}s_y\times{}in_{ch}$. 
Standard convolutions have a high computational cost, depending on the size of the kernel and the size of the input. 
To optimize this process, depth-separable convolution was introduced. 
Specifically, depth-separable convolution involves two steps:
\begin{enumerate}
\item Performing a point-wise convolution with $1\times{}1$ filters, resulting in a feature map with a depth of $out_{ch}$. 
\item Conducting a spatial convolution with $s_x\times{}s_y$ filters in the $x$ and $y$ dimensions. 
\end{enumerate}
By stacking these two operations without intermediate non-linear activation, the output shape remains the same as that of a regular convolution, but with significantly fewer parameters. 

This technique is utilized in models like MobileNet~\cite{howard2017mobilenets}, and MobileNetV2~\cite{sandler2018mobilenetv2}, designed for mobile and embedded devices. 
Using depth-wise separable layers instead of regular convolutions, MobileNet reduces the number of parameters and multiply-add operations, enabling efficient deployment on mobile devices for computer vision tasks. 

Furthermore, in~\cite{luo2022tiny} Tiny-Sepformer is presented, a tiny time-domain transformer network, that uses Convolution-Attention (CA) block into the masking network, in order to split the layer into convolution path and attention path parallelly. 
In particular, to further reduce the computation, the convolution part of CA is a 1D depthwise separable convolution. 

Finally, in~\cite{chollet2017xception} Xception is presented, an interpretation that considers Inception modules in convolutional neural networks as an intermediate step between regular convolution and the depthwise separable convolution operation (a depthwise convolution followed by a pointwise convolution). 

\subsubsection{Attention Mechanism}
This technique, while commonly associated with notable contributions to machine translation tasks, has been adapted and adopted for a wide range of applications~\cite{brauwers2021general}. 
Its fundamental purpose remains unchanged: allow the model to focus on relevant parts of the input while generating outputs. 
This enables the model to selectively attend to different regions or features, thereby facilitating the extraction of salient information from complex and high-dimensional data. 
Beyond translation tasks, the attention mechanism has been successfully employed in natural language processing~\cite{galassi2020attention}, image captioning~\cite{huang2019attention}, speech recognition~\cite{chorowski2015attention}, and more. 
By incorporating attention into these tasks, models can effectively handle long-range dependencies, capture context-specific information, and improve overall accuracy and robustness. 

The beauty of the attention mechanism lies in its ability to assign importance to different parts of the input dynamically, based on their relevance to the current context. 
This adaptability allows models to prioritize relevant information and disregard noise or irrelevant details, resulting in more precise and context-aware predictions. 
Hence, attention can be particularly useful in \tinyml{} applications, where resource-constrained devices require efficient and compact models. In~\cite{wong2020attendnets}, the authors introduce AttendNets, a deep self-attention architecture based on visual attention condensers, to deploy on-device visual perception tasks like image recognition.

\subsection{Learning Algorithms}
\begin{table}[t!]
    \centering
    \caption{Popularity and their respective papers over the years of learning algorithms based on referenced research contributions.}
    \begin{scriptsize}
    \input{tables/t_learning_algorithms}
    \end{scriptsize}
    \label{tab:t_learning_algorithms}
\end{table}
Among the many taxonomies covering the whole discipline of ML, we focus first on the standard paradigms of \textit{\textbf{supervised}}, \textit{\textbf{unsupervised}}, \textit{\textbf{Self-Supervised (SSL)}}, and \textit{\textbf{deep reinforcement learning}}. 
Moreover, for the particular case of optimal usage of resources, we review \textit{\textbf{weakly-supervised learning}}, \textit{\textbf{meta-learning}} and \textit{\textbf{continual learning}} techniques, detailing how these are useful for \tinyml{}. 
Also in this case, Table~\ref{tab:t_learning_algorithms} provides an overview of the popularity of these techniques within the \tinyml{} field, considering the referenced research contributions. 

\subsubsection{Supervised Learning}
In this paradigm, a model learns from labeled training data to make predictions. 
It involves training a model on input-output pairs, where the input data is fed into the model, and the corresponding desired output or label is provided. 
This training process allows for robust models but at the cost of requiring a large number of annotated data. 
Supervised learning, in general, guarantees good performances when a lot of training data is available, which usually also requires big models with a high number of parameters that are capable of learning the patterns present in the data. 

In this sense, there are \tinyml{} approaches that perform supervised learning on big models and then distill the knowledge on small models leveraging on the knowledge learned from the big supervised models. 
This can be performed by compressing a previously trained model (teacher) to a smaller one (student) as in~\cite{abbasi2020compress}, in a process that is similar to the knowledge distillation process (as explained in Section~\ref{sub:kd}), or by learning directly from a big model leveraging parallel computation, as proposed by~\cite{de2021robustifying}. 

As stated before, supervised learning requires a lot of annotated data to be effective. 
Despite that, some \tinyml{} works try to compensate for the lack of annotation on the data. 
In this regard, in~\cite{hussein2023sensorgan} the authors implement a method for generating missing samples during training in the context of human action recognition, where missing samples could lead to inaccurate classifications. 
In~\cite{yamin2023uncertainty}, they propose to take into account the uncertainty of future samples in a power consumption management implant. 
This is performed by trying to capture the underlying seasonal and daily changes from some annotated data, and then forecast the uncertainty of future energy consumption. 

\subsubsection{Unsupervised Learning} 
In this paradigm, a model learns from unlabeled data, striving to discover patterns, structures, or relationships in the data. 

Anomaly detection is one of the most common use cases for employing the unsupervised learning approach. 
This makes anomaly detection particularly interesting for \tinyml{}, given the need to process raw data streams in their early stages, as close to their origins as possible. 
Specifically, the industrial environment is where a majority of research and development related to anomaly detection is concentrated~\cite{abbasi2021outliernets}. 
This is due to the unique challenges encountered in such environments, including limited or unreliable communication with the cloud, uninterrupted connectivity, and potential obstacles to accessing systems. 
In this sense, \tinyml{} becomes of necessary importance. 

Besides the industrial environment, other real-world scenarios are interested in unsupervised anomaly detection, such as physiological disorders~\cite{goyal2020drocc} and climate conditions~\cite{kayan2021anoml}, in which classic machine and deep learning models are proposed. 

Specifically regarding deep learning approaches, in~\cite{lord2021mechanical} a model based on autoencoders has been engineered to be executed on a microcontroller for detecting anomalies of top-load industrial washing machines. 
The model has been ported to an Arduino Nano microcontroller, achieving high accuracy and recall performances, with remarkably low power usage. 

Finally, related to the empirical data analysis approach, in~\cite{andrade2021unsupervised}, the authors propose an unsupervised \tinyml{} approach to detect anomalies on roads, based on the concept of Typicality and Eccentricity of Data (TEDA). 
Similar work is presented in~\cite{andrade2022tinyml}, where the focus lies on monitoring the release of greenhouse gases from urban vehicles. 
Specifically, a \tinyml{} unsupervised methodology is employed to quantify $\text{CO}_2$ emissions for the evaluation of air quality within urban environments. 

\subsubsection{Self-Supervised Learning (SSL)}
Supervised learning is currently facing a bottleneck due to its significant dependence on expensive manual labeling, leading to issues such as generalization errors and spurious correlations~\cite{liu2021self}. 
SSL has emerged as a highly promising technique to address the aforementioned challenges, offering a solution that eliminates the need for costly and expensive manual annotations~\cite{rani2023self}. 

In SSL, the model is trained to predict certain aspects of the input data without relying on external annotations. 
This can be achieved by creating surrogate tasks such as, in the case of image-based tasks, predicting missing parts of an image, reconstructing an image from a corrupted version, and predicting the relative position of image patches. 
In this manner, the model is encouraged to capture the underlying structure and semantics of the data. 

The synergy between SSL and anomaly detection is evident: by reconstructing or predicting parts of the data without explicit labels, the model learns to capture the inherent structure of the majority class, making it more sensitive to deviations and anomalies. 
Furthermore, is particularly advantageous when labeled anomaly data is scarce or expensive to obtain, as it enables the model to generalize better and identify anomalies effectively in diverse and complex datasets. 
Since anomaly detection faces a scarcity of labeled anomaly examples, SSL leverages unlabeled data to learn features that capture the underlying structure of the data, effectively utilizing the abundance of unlabeled data available. 

Specifically, in~\cite{abououf2022self} an SSL method for anomaly detection in IoT devices is proposed. 
Their method is based on a multivariate Long Short-Term Memory (LSTM) autoencoder. 
The self-supervision is performed by detecting data that significantly deviates from the learned distribution, and using them as anomalous data to enhance the detection of such anomaly types. 

\subsubsection{Deep Reinforcement Learning}
Reinforcement Learning (RL) is a ML technique that trains a model (usually an agent) to take actions (policies) based on the input. 
The way the model learns is usually based on rewards assigned to a good policy that the model needs to maximize in the form of a Markov Decision Process (MDP)~\cite{kaelbling1996reinforcement}. 
Integrated with deep learning, \ie{}, deep reinforcement learning, the optimal policies are obtained. 
This is useful because in real-world applications the space state is high-dimensional and the use of traditional RL algorithms is not effective~\cite{arulkumaran2017deep}. 
Specifically, in the case of \tinyml{}, the challenge is to embed the neural networks implemented with deep reinforcement learning approaches on small, constrained devices such as MCUs. 

Transferring trained deep reinforcement learning models on constrained devices is possible using several general-purpose techniques usually designed to alleviate the system resource bottlenecks, as proposed by~\cite{svoboda2020resource} in their framework suite, making feasible deep reinforcement learning for \tinyml{} platforms. 
Moreover, in~\cite{szydlo2022tinyrl}, the authors propose the framework TinyRL, to transfer the deep reinforcement learning knowledge into resource-limited devices. 

Cheap off-the-shelf MCU devices are particularly interesting for deep reinforcement learning as they are widely adopted in robotics. 
Deploying a DRL model on an MCU-powered intelligent agent for autonomous driving, for instance, is what the authors of~\cite{kong2022deep} propose to achieve. 
They present a deterministic policy gradient algorithm that takes into consideration the computation energy and caching costs jointly. 
This significantly reduces the energy cost of the final model. 
Moreover, in the efforts of allowing deep reinforcement learning on MCU, \cite{pau2023end} propose to train a tiny CNN that can be easily deployed on an MCU, with the aim of solving a physical, electrically actuated tilting maze with repositionable walls. 

\subsubsection{Weakly Supervised Learning}
This paradigm lies between supervised and unsupervised learning. 
It involves training a model using partially labeled or noisy labeled data, where only limited or incomplete supervision is available. 
Since collecting large amounts of accurately labeled data can be expensive and time-consuming, weakly supervised learning allows for training models with fewer labeled samples, reducing the labeling costs associated with data collection. 
This cost-efficiency is clearly advantageous for \tinyml{} applications, where resource constraints often limit the availability of labeled training data. 

For example, in~\cite{barbariol2022tiws} the authors propose a weakly supervised learning solution for achieving improved anomaly detection performances. 
In particular, the training phase of the model is improved by some labels in the dataset: in fact, a part of the dataset is labeled, playing the role of ``domain expert,'' which allows weakly supervised learning. 
The ML model used is an Isolation Forest, and these labels are used to remove unnecessary trees and keep the most informative ones, \ie{}, those that give the best results). 

In~\cite{antonini2023adaptable} is presented another \tinyml{}-based system for anomaly detection in industrial environments. 
In this case, an ensemble of ML classifiers detects if a sample is anomalous or not. 
This allows the system to be scalable \wrt{} the size of the ensemble, with a predictable impact on the memory footprint and delay in inference mode. 

\subsubsection{Meta Learning}
Unlike traditional methods that solve tasks independently using a fixed learning algorithm, meta-learning enhances the learning algorithm itself based on experiences from multiple learning episodes~\cite{hospedales2021meta}. 
While various perspectives on meta-learning exist, our focus here is on the optimization approach known as neural-network meta-learning, which is particularly relevant for \tinyml{} applications. 
The neural-network meta-learning design should take into account three independent axes that represent the current meta-learning landscape: \textit{\textbf{meta-representation}}, \textit{\textbf{meta-optimizer}}, and \textit{\textbf{meta-objective}}. 

In particular, the concept of \textit{meta-representation}~\cite{finn2017model} involves learning a high-level representation that captures the commonalities and patterns across different tasks. 
This meta-representation serves as a knowledge base from which the model can quickly adapt to new tasks with limited data. 
For example, in~\cite{liu2023metaldc} MetaLDC is proposed, a system that meta-trains ultra-efficient low-dimensional computing classifiers to enable fast adaptation on tiny devices with minimal computational costs. 
Specifically, during the meta-training stage, MetaLDC meta-trains a representation offline by explicitly taking into account that the final (binary) class layer will be fine-tuned for fast adaptation for unseen tasks on tiny devices; during the meta-testing stage, MetaLDC uses closed-form gradients of the loss function to enable fast adaptation of the class layer. 

On the other hand, the \textit{meta-optimizer}~\cite{belkhale2021model} learns to optimize the model's parameters in a way that facilitates fast adaptation and generalization across tasks. 
Specifically, the meta-optimizer tunes the learning algorithm itself, enabling it to efficiently update the model based on task-specific information. 
TinyReptile, a simple but efficient meta-optimizer-based algorithm to collaboratively learn a solid initialization for a neural network across tiny devices, is presented in~\cite{ren2023tinyreptile}. 

Lastly, the \textit{meta-objective}~\cite{cho2021camera} guides the meta-learning process by defining a criterion for evaluating the performance of the meta-learner. 
It provides a signal for learning to adapt and generalize, encouraging the acquisition of task-agnostic knowledge that can be applied to new tasks. 
To the best of our knowledge, the two closest works in this area are~\cite{gao2021pruning}, in which the authors propose an Adaptation-aware Network Pruning (ANP), a novel pruning scheme that works with existing meta-learning methods for a compact network capable of fast adaptation, and~\cite{gao2022finding}, in which it is shown that the application of Lottery Ticket Hypothesis (LTH) to meta-learning enables the adaptation of meta-trained networks on various IoT devices. 

\subsubsection{Continual Learning}
This research field aims at developing algorithms that enable models to continuously learn from new data while preserving previously acquired knowledge. 
This is essential because conventional ML models typically struggle to learn from new data while retaining previously acquired knowledge, often leading to catastrophic forgetting~\cite{kirkpatrick2017overcoming}. 

Recently, there has been a significant development in the field, with several promising algorithms and architectures being proposed and showing improved performance in various continual learning benchmarks. 
These approaches can be classified into the following three categories: \textit{\textbf{regularization-based}}, \textit{\textbf{replay-based}}, and \textit{\textbf{dynamic architectures}}. 

Specifically, \textit{regularization-based} methods introduce regularization terms into the loss function to encourage the model to maintain its prior knowledge while learning new tasks. 
On the other hand, \textit{replay-based} methods involve storing past data and replaying it during training to prevent forgetting. 
Finally, \textit{dynamic architectures}, adjust their capacity to accommodate new information. 

The significance of this research area has increased due to the surging demand for \tinyml{} models in several applications, including healthcare, wearables, and IoT devices. 
Recent research in this domain has been focused on the development of efficient algorithms that can manage the restrictions of \tinyml{} devices, such as limited memory and processing power. 
One of the promising methods in a \tinyml{} scenario is to use regularization-based approaches that add penalties to the loss function to prevent overfitting and catastrophic forgetting. 
Another effective approach is to use dynamic architectures that can adapt their structure to accommodate new tasks. 
For instance, in~\cite{avi2022incremental} a regularization-based approach for an IoT scenario is presented, in which MCUs are exploited as edge devices for data processing considering two tasks: gesture recognition based on accelerometer data and image classification. 

Replay-based methods, on the other hand, are well adopted in real-world scenarios as the general replay approach is very intuitive. In~\cite{ravaglia2021tinyml}, the authors leverage the quantization of the frozen stage of the model, allowing for 8-bit execution, and replays in the latent space to reduce their memory cost with minimal impact on accuracy. 
The results show that by combining these techniques, continual learning can be achieved in practice using less than 64MB of memory an amount compatible with embedding in \tinyml{} devices. 
In~\cite{sudharsan2021train++}, the authors propose Train++, an incremental replay-based training algorithm that trains ML models locally at the device level (\eg{}, on MCUs) using the full n-samples of high-dimensional data. 
Train++ enables resource-constrained MCU-based IoT edge devices to locally build their own knowledge base on the fly using the live data, thus creating smart self-learning and autonomous problem-solving devices.  
The authors of~\cite{pavan2023tybox} propose TyBox, a toolbox for the automatic design of on-device \tinyml{} classification models, with the idea of automatically generating the ``incremental'' version of an initial (static) pre-trained model using replays. 

Lastly, regarding the dynamic architectures approaches for continual learning, in~\cite{cai2020tinytl} a pioneering contribution in the form of Tiny-Transfer-Learning (TinyTL) is presented. 
In their work, the authors propose a novel approach that achieves memory efficiency by selectively freezing the weights of the network while solely focusing on learning the bias modules, thereby obviating the need to store intermediate activations. 
To ensure the adaptability of the model, a new memory-efficient bias module, referred to as the lite residual module, is introduced. 
Through extensive experimentation, it is demonstrated that TinyTL yields substantial memory savings with minimal sacrifice in accuracy compared to the conventional fine-tuning approach applied to the entire network. 
Finally, in~\cite{ren2021tinyol} the authors propose TinyOL (TinyML with Online Learning), which enables incremental on-device training with streaming data. 

%% file: tables/t_model_optimization.tex
\begin{tabular}{lcccc}
\toprule
\textbf{Technique} & \textbf{Popularity} & \textbf{Pre '20} & \textbf{'20 - '22} & \textbf{'23} \\ 
\midrule
Pruning & H & \cite{hu2016network,anwar2017structured,zhu2017prune,yu2017scalpel,liu2018rethinking} & 
          \cite{blalock2020state,de2022depth,vadera2022methods}                                     &  
          \cite{sun2023case,hashir2023tinyml}                                                       \\
Quantization & H & \cite{cai2017deep,jacob2018quantization,krishnamoorthi2018quantizing,mishchenko2019low}  &
               \cite{nagel2020up,wang2020towards,gholami2022survey,nagel2022overcoming,zhuo2022empirical}   & 
               \cite{moosmann2023tinyissimoyolo,lu2023enhancing,alajlan2023ddd}                             \\
Knowledge distillation & H &                        &
\cite{yun2020regularizing,zhao2020highlight,zhang2020improve,korber2021tiny,ukil2021resource,gou2021knowledge,zhang2021adversarial,cheng2021relation,dai2021general,al2022implementation,brutti2022optimizing}       & \\
HPO & L                                                               &                                   
\cite{bergstra2011algorithms,bergstra2012random,wu2019hyperparameter} & 
\cite{sandha2021enabling}                                             \\
\bottomrule
\end{tabular}

%% file: tables/t_model_design.tex
\begin{tabular}{lcccc}
\toprule
\textbf{Technique} & \textbf{Popularity} & \textbf{Pre '20} & \textbf{'20 - '22} & \textbf{'23} \\ 
\midrule
NAS & H & & 
\cite{mendis2021intermittent,banbury2021micronets,liberis2021munas,ren2021comprehensive,baymurzina2021review,liu2021survey} & \cite{njor2023data,pau2023quantitative,garavagno2023hardware}                                                               \\
RAF & L & \cite{molina2019pad} & \cite{apicella2021survey,trimmel2022era} & \\
Depth-separable & M                                                    & 
\cite{howard2017mobilenets,chollet2017xception,sandler2018mobilenetv2} & 
\cite{luo2022tiny}                                                     & \\
Attention & M                                    & 
\cite{chorowski2015attention,huang2019attention} & \cite{galassi2020attention,wong2020attendnets,brauwers2021general,burrello2021microcontroller,mehta2021mobilevit} & \\
\bottomrule
\end{tabular}

%% file: tables/t_learning_algorithms.tex
\begin{tabular}{lcccc}
\toprule
\textbf{Technique} & \textbf{Popularity} & \textbf{Pre '20} & \textbf{'20 - '22} & \textbf{'23} \\ 
\midrule
Supervised & M & & \cite{abbasi2020compress,de2021robustifying} & \cite{hussein2023sensorgan,yamin2023uncertainty}                             \\
Unsupervised & H & & \cite{goyal2020drocc,abbasi2021outliernets,kayan2021anoml,lord2021mechanical,andrade2021unsupervised,andrade2022tinyml} & \\
SSL & L & & \cite{liu2021self,abououf2022self} & \cite{rani2023self}                                                      \\
Deep RL & M & \cite{kaelbling1996reinforcement} & \cite{svoboda2020resource,szydlo2022tinyrl,kong2022deep} & \cite{pau2023end}                 \\
Weakly supervised & L & & \cite{barbariol2022tiws} & \cite{antonini2023adaptable}                                                              \\
Meta learning & M & \cite{finn2017model}                                                & 
\cite{belkhale2021model,cho2021camera,gao2021pruning,gao2022finding} & 
\cite{liu2023metaldc,ren2023tinyreptile}                                                \\
Continual learning & H & 
\cite{kirkpatrick2017overcoming} & \cite{cai2020tinytl,ren2021tinyol,ravaglia2021tinyml,sudharsan2021train++,avi2022incremental} & \cite{pavan2023tybox}   \\
\bottomrule
\end{tabular}

%% file: srcs/6_devices_and_tools.tex
\section{TinyML Devices and Tools}
\label{cha:6_devices_and_tools}

\begin{table*}[t!]
    \centering
    \caption{Comparison between the different hardware for a \tinyml{}-based system.}
    \input{tables/t_hardware}
    \label{tab:t_hardware}
\end{table*}
\tinyml{} heavily depends on hardware devices to enable efficient training and inference for its applications. 
In particular, \textit{\textbf{Central Processing Units (CPUs)}}, \textit{\textbf{Graphics Processing Units (GPUs)}}, \textit{\textbf{FPGAs}}, and \textit{\textbf{Tensor Processing Units (TPUs)}} are essential components for the functioning of \tinyml{}-based systems. 
Table~\ref{tab:t_hardware} presents a summary of a comparison between these different hardware devices, outlining their respective advantages and disadvantages. 
The following sections present a summary of a comparison between these different hardware devices, outlining their respective advantages and disadvantages.

\subsection{Central Processing Unit (CPU)}
The primary objective of \tinyml{} is to optimize ML workloads in a way that allows them to be executed on microcontrollers with extremely low power consumption, often just a few milliwatts. 
Microcontrollers, particularly the Arm Cortex-M family, serve as an ideal platform for implementing ML due to their widespread usage~\cite{suda2019machine}. 
Additionally, their minimal power requirements make them suitable for deployment in environments where replacing batteries is challenging or inconvenient~\cite{lai2018cmsis}. 

However, despite their advantages, microcontrollers exhibit several drawbacks. 
Their wide-ranging applicability leads to the inclusion of unnecessary operations and logic checks, which might degrade computational performance. 
In addition, this fails to fully exploit the potential parallelism offered by deep learning algorithms.

\subsection{Graphics Processing Unit (GPU)}
Originally designed for accelerating computer graphics, GPUs differ from CPUs in their composition. 
While CPUs consist of a few Arithmetic Logic Units (ALUs) optimized for sequential processing, GPUs are equipped with thousands of ALUs that enable parallel execution of numerous simple operations. 
This parallel architecture makes GPUs highly suitable for ML tasks since they can rapidly perform a large number of parallel computations. 
For example, ML algorithms often involve extensive matrix and vector operations, which can be efficiently parallelized and executed on GPUs, as we can see in~\cite{krizhevsky2012imagenet}. 

In recent years, Nvidia has introduced multiple generations of GPU microarchitectures, such as the Nvidia Jetson family, with a growing emphasis on enhancing deep learning performance. 
Additionally, Nvidia has introduced Tensor Cores~\cite{markidis2018nvidia}, specialized execution units within their GPUs specifically designed for deep learning applications. 

Furthermore, GPUs are specifically designed to efficiently handle large datasets and facilitate rapid data transfer between the main system memory and processing units, a significant attribute since ML typically operates on extensive real-time data~\cite{wang2019benchmarking}. 
Hence, the combination of parallel computing capabilities, a significant number of cores, and high-bandwidth memory access collectively establish GPU microarchitecture as a good choice for \tinyml{}-based applications.

\subsection{Field-Programmable Gate Array (FPGA)}
An FPGA offers a high-performance, efficient, and scalable solution for handling the intricate mathematical computations demanded by ML~\cite{shawahna2018fpga}. 

The fundamental building block of an FPGA's architecture is the fabric layer, comprising Configurable Logic Blocks (CLBs) and programmable interconnects. 
CLBs can be flexibly configured by users to perform various digital functions, including the complex mathematical operations essential for ML algorithms. 
Through programmable interconnects, CLBs can be interconnected in different configurations, enabling customization to suit diverse ML applications~\cite{nasri2009design}. 

Aside from the fabric layer, an FPGA designed for ML often incorporates additional specialized hardware blocks, such as Digital Signal Processing (DSP) blocks and high-performance memory blocks. 
DSP blocks are utilized to enhance the execution speed of intricate mathematical operations like convolutions and dot products, which are commonly employed in ML models. 
High-performance memory blocks also facilitate rapid access to the extensive datasets~\cite{wang2018survey}. 

To summarize, an FPGA aims to deliver high performance, efficiency, and scalability, catering to the complex computational requirements of ML tasks. 
Consequently, FPGAs emerge as an ideal choice for a broad spectrum of applications in the \tinyml{} domain.

\subsection{Tensor Processing Unit (TPU)}
A specialized processor, known as the TPU, has been developed by Google explicitly for ML tasks, with a specific emphasis on tensor operations~\cite{jouppi2017datacenter}. 
TPUs consist of several part, including a high-bandwidth memory system, a systolic array of processing units, and an interconnected network that facilitates communication between these components. 

The systolic array represents the core of the TPU, responsible for executing tensor operations. 
It consists of numerous processing elements arranged in a two-dimensional grid, with each element interconnected to its neighboring ones. 
This arrangement enables efficient communication between processing elements, facilitating the parallel execution of complex tensor operations. 
The high-bandwidth memory system ensures swift access to data necessary for tensor operations. 
Finally, the interconnect network links the TPU with other system components such as the host processor and other TPUs, promoting efficient communication and coordination among them. 
Google has released Edge TPUs using the Coral platform in various form factors, ranging from a Raspberry-Pi-like Dev Board to stand-alone solderable modules~\cite{seshadri2022evaluation}. 

In summary, the architecture of an Edge TPU is specifically designed to provide high performance, efficiency, and scalability in handling tensor operations, making it an excellent choice for \tinyml{} applications.

\subsection{Software Tools}
\label{sec:6_e_sw_tools}
As the demand for implementing ML on various hardware devices continues to grow, the software layer emerges as one of the essential components in the development of \tinyml{}-based systems. 
To date, prevalent frameworks heavily rely on vendor-specific operator libraries, demonstrating the significant potential for driving advancements in \tinyml{} research. 
Below, we provide an overview of the main frameworks utilized in this domain:
\begin{itemize}
\item \textbf{TensorFlow Lite Micro}~\cite{david2021tensorflow}: is an open-source framework that empowers microcontrollers and similar devices with limited memory capacity to execute ML models. 
It operates efficiently without relying on an operating system, standard C (or C++ libraries), or dynamic memory allocation. 
Developed in C++11, this framework necessitates a 32-bit platform and exhibits compatibility with most Arm Cortex-M Series processors. 

\item \textbf{uTensor}~\cite{utensor}: is a remarkably lightweight, open-source framework for ML inference. 
It is built upon TensorFlow and meticulously optimized for Arm targets. 
By converting ML models into readable and self-contained C++ source files, uTensor greatly simplifies integration with embedded projects. 

\item \textbf{Edge Impulse}~\cite{janapa2023edge}: is a service that facilitates the development of \tinyml{} models specifically tailored for edge devices. 
The training process takes place on a cloud platform, and the resulting trained model can be easily exported to an edge device. 
Additionally, Edge Impulse simplifies the collection of actual sensor data, enables live signal processing from raw data to neural networks, and streamlines testing procedures. 

\item \textbf{Embedded Learning Library}~\cite{embeddedlearninglibrary}: the Microsoft Embedded Learning Library (ELL) empowers users to design and implement intelligent ML models on resource-constrained platforms.  
Conceptually, the ELL can be seen as a cross-compiler for intelligence embedding, where the compiler operates on the laptop and generates machine code that can be executed on the embedded device. 

\item \textbf{X-CUBE-AI}~\cite{xcubeai}: is an STM32Cube expansion package. 
Allows an automatic conversion of pre-trained artificial intelligence algorithms, including neural networks and classical ML models, for STM products. 
It also integrates an optimized library for STM32 ARM Cortex M-based boards. 

\item \textbf{uTVM}~\cite{chen2018tvm}: is a compiler that offers graph-level and operator-level optimizations, enabling deep learning workloads to achieve performance portability across a wide range of hardware back-ends. 
It addresses optimization challenges specific to deep learning, including high-level operator fusion, mapping to various hardware primitives, and effectively mitigating memory latency. 

\item \textbf{MinUn}~\cite{jaiswal2023minun}: is a framework jointly developed by Microsoft Research in India, ETH Zurich, and UC Berkeley, specifically designed for \tinyml{} applications. It presents a comprehensive solution to three critical sub-problems. 
Firstly, it addresses the challenge of utilizing number representations that approximate 32-bit floating point numbers using fewer bits, without compromising accuracy. 
Secondly, it offers heuristic techniques to optimize bandwidth assignment, ensuring minimal memory usage while preserving accuracy, and lastly, it tackles the issue of memory management on devices with limited resources, mitigating potential problems related to memory fragmentation. 
\end{itemize}

%% file: tables/t_hardware.tex
\begin{tabular}{lll}
\toprule 
\textbf{Hardware} & \textbf{Advantage} & \textbf{Disavantage} \\
\midrule
CPU        & Fit for general purpose, high memory capacity                  & Low parallelism, low throughput performance      \\
GPU        & High throughput performance, a good fit for SOTA architectures & Expensive, energy-hungry                         \\
FPGA       & Energy efficient, flexible                                     & Extremely difficult to use, lack of libraries    \\
TPU        & Potential to significantly boost inference performance         & Expensive, hard to develop                       \\
\bottomrule
\end{tabular}

%% file: srcs/7_discussion.tex
\section{Discussion}
\label{cha:7_discussion}

In the recent five years, as shown by our search strategy, there has been a notable surge in studies investigating \tinyml{} methods, optimizations, and applications. 
This trend reflects the growing recognition of the importance of real-time solutions for many complex and safety-critical real-world applications. 
In this paper, we present a comprehensive analysis from the ML point of view of \tinyml{}. 
Our aim is to provide not just an updated guide on the current state-of-the-art, but also to pinpoint areas that have yet to be explored. 
By doing so, we hope to lay the foundation for future research and investigations in this field. 

From the proposed taxonomy in Figure~\ref{fig:efficient_DL}, the area of model optimization, based on referenced research contributions, is the one that has received the most extensive exploration. 
Indeed, within \tinyml{}, we come across several cutting-edge works that explore techniques such as pruning~\cite{de2022depth}, quantization~\cite{zhuo2022empirical}, and knowledge distillation~\cite{al2022implementation}. 
On the contrary, the scarcity of research focusing on HPO~\cite{sandha2021enabling} can be attributed to its complexity, the lack of awareness about the importance of HPO and its potential to significantly enhance the performance of ML models, and the resource requirements, which can be a limiting factor for researchers with restricted access to high-performance computing infrastructure. 
Table~\ref{tab:t_model_optimization} summarizes this inquiry. 

In terms of the model design area, most of the work is focused on NAS~\cite{banbury2021micronets}. 
Secondly, it's worth noting that another significant portion of the existing research in this area is related to attention mechanisms~\cite{burrello2021microcontroller,mehta2021mobilevit} and depth-separable convolutions since these techniques enhance model efficiency, accuracy, and real-time inference capabilities, making them essential for resource-constrained edge devices. 
Therefore, we expect that research in these areas will continue to grow significantly in the coming years. 
However, despite the numerous advancements made in ML through the application of RAFs or VeLO, no such method has been found in any works related to \tinyml{}. 
Regarding this, we believe that pursuing research in this direction could potentially lead to further enhancements in the quality of the produced models. 
In this case, Table~\ref{tab:t_model_design} resumes this information. 

Related to the learning algorithms, we have encountered a significant body of research focused on unsupervised learning and continual learning. 
Additionally, considerable efforts have been made in the domains of supervised learning, meta-learning, and deep reinforcement learning. 
However, we note that the fields of self-supervised learning and weakly-supervised learning still require effort before they can be widely used in \tinyml{}-based work. 
Hence, we strongly believe that directing research efforts toward these two areas holds immense potential for significant advancements in the years to come. 
Table~\ref{tab:t_learning_algorithms} sums up this direction. 

In the field of \tinyml{} applications, most efforts focus on addressing the challenges of anomaly detection. 
This is consistent with the previous statement emphasizing the use of unsupervised and continual learning strategies. 
It is worth noting that within the \tinyml{} landscape, the majority of methodologies used to tackle various tasks lean toward deep learning paradigms. 
Non-deep learning algorithms are used sporadically in this context, with notable exceptions being approaches based on the TEDA framework, as demonstrated in~\cite{andrade2021unsupervised}. 

Regarding hardware choices, there isn't a one-size-fits-all solution among CPUs, GPUs, FPGAs, and TPUs. 
Each of them comes with its own set of advantages and disadvantages, as summarized in Table\ref{tab:t_hardware}. 
Therefore, selecting the most suitable hardware for a specific application is essential. 
However, we expect that the co-design approach will significantly focus further advancements in this research field since the early work seems to be extremely promising, making it the primary direction of future developments. 

Despite the promising applications and growing scientific literature in the field of \tinyml{}, further research is needed to fully comprehend its advantages and limitations. 
In this context, we draw other additional unresolved issues that require dedicated research to drive future advancements in the field. 
\begin{itemize}
\item \textbf{Benchmarking}: The lack of a recognized benchmark, due to the challenges posed by low power, limited memory, hardware heterogeneity, and software heterogeneity, is an important impediment that may hamper \tinyml{} services~\cite{banbury2020benchmarking}. 
In this context, the IoT community has shown an increasing interest in benchmarking as a way to scientifically compare the performance of various \tinyml{} solutions, both for training benchmark~\cite{mattson2020mlperft}, for inference benchmark~\cite{reddi2020mlperf}, and specifically for \tinyml{} systems~\cite{banbury2021mlperf}. 

\item \textbf{Memory Constraints}: The insatiable demand for computation and high accuracy has continued to push the innovations in ML algorithms, but the extremely small size of SRAM and flash memory makes the task of deep learning on edge devices even today very challenging. 

\item \textbf{Data-driven engineering}: It's critical to understand data quality thoroughly because relying solely on accuracy can be misleading when predicting model behavior. 
To accomplish this, we will need a large amount of relevant real-world data. 
This information will assist us in identifying specific instances where the model fails to detect or behaves incorrectly. 
Furthermore, post-processing techniques will be required to improve the model's performance in these areas. 
In essence, we need tools and processes that prioritize ``data excellence'' in order to assess data quality comprehensively. 

\item \textbf{Lack of accepted models}: Currently, many deep learning models are widely accepted for conventional infrastructure. 
For example, MobileNet is used as the baseline for benchmarking deep neural networks in mobile edge computing devices. 
However, for example, no such popular model can be adopted for the \tinyml{} on the MCUs ecosystem. 

\item \textbf{Lack of public datasets}: Despite some datasets specifically designed for \tinyml{} being available (such as for on-device online training~\cite{sudharsan2021train++}), to date, \tinyml{} is mainly concerned with sensor processing in general, so the question that emerges is\dots{} \textit{``What's the ImageNet~\cite{deng2009imagenet} of \tinyml{}''?} 
\end{itemize}

%% file: srcs/8_conclusion.tex
\section{Conclusion}
\label{cha:8_conlusion}

The prodigious amount of research invested over the past decades in improving embedded technologies to enable the use of real-time solutions for many complex and safety-critical applications led to the birth of \tinyml{} (Section~\ref{cha:1_intro}). 
As summarized in Figure~\ref{fig:PRISMA-on-tinyML}, this paper presents a systematic review of \tinyml{} from January 2018 to September 2023 (Section~\ref{cha:3_selection_criteria}). 
For the first time ever, we formalize the three different workflows to implement a \tinyml{}-based system (Section~\ref{cha:4_tinyML_pipelines}). 
As an additional and distinct contribution, this survey places a strong emphasis on the ML perspective. 
It not only presents the most current \tinyml{} frameworks but also recommends recent variations and advancements in ML technologies that \tinyml{} practitioners may consider exploring to enhance the state-of-the-art capabilities (Section~\ref{cha:5_efficient_dl}). 
In Section~\ref{cha:6_devices_and_tools} we examine the advantages and disadvantages of different hardware devices that can be used to develop \tinyml{}-based applications. 
Finally, in Section~\ref{cha:7_discussion} we highlight the fields that hold the most promise for further research in the upcoming years. 
Additionally, we provide a list of unresolved problems that need to be addressed in order to propel the field forward.